# Theory-Guided Deception Detection:

# A RAG-Based Artificial Intelligence Exploration

David M. Markowitz[1] & Timothy R. Levine[2]

[1] Department of Communication, Michigan State University, East Lansing, MI 48824

[2] Department of Communication, University of Oklahoma, Norman, OK 73019

## Abstract

The current work developed seven Retrieval-Augmented Generation (RAG) models based on leading deception theories and compared how deception judgments were made relative to baseline models. Across 700 statements drawn from five published deception datasets, four large language models (*gpt-4o*, *claude-sonnet-4-6*, *ollama/llama3*, *deepseek-v4-flash*), and two run-types (RAG vs. baseline), a total of 39,200 deception judgments were rendered. Detection accuracies were consistent with typical human accuracies and not statistically different across RAG (54.5%) and baseline models (54.6%). RAG-based models (57.0%) were less truth-biased than baseline models (59.7%), but the effect size was quite small. Theoretical perspective mattered little for accuracy yet mattered substantially for response bias, which ranged from highly lie-biased (the verifiability approach, 32.2%) to highly truth-biased (truth-default theory, 88.1%). Content effects and model effects further moderated the results. Theory-guided AI judgments are unreliable with current parameters, yet they might show promise with additional datasets, model testing, and theory-to-data matching.

**Public Significance Statement:** How well can theory-guided AI models detect deception? Using seven deception theories, we examined this question across different published datasets and large language models. While detection accuracy was roughly equivalent between theory-guided and baseline models, theory-guided models were less truth-biased. Deception detection with AI is still in its infancy.

**Theory-Guided Deception Detection:**

**A RAG-Based Artificial Intelligence Exploration**

Scholars and practitioners have long imagined a future where technology can aid in deception detection. For example, an immense number of resources have been dedicated toward examining new physiological (e.g., the polygraph), behavioral (e.g., facial expression analysis and eye tracking), neuroimaging (e.g., fMRI), and computational (e.g., natural language processing) approaches to distinguishing truthful from deceptive communication — many of which have yielded mixed or middling effects. As a new technology emerges, there is a tendency to view it as a potential "silver bullet" for deception detection. Indeed, there is a utopian undercurrent to this view. Given the unreliability of human judgments in deception detection (Bond & DePaulo, 2006; Hartwig & Bond, 2014; Hartwig & Bond Jr, 2011), some researchers hold out hope for a new tool or set of cues to betray veracity with enough signal that they will catch liars in the act, or at least make them think twice before communicating falsely to another person. In other words, more technological sophistication increases the probability that deception scholars will become interested in another tool to detect deceit. Unfortunately, decades of promise have been met with the realities of false promises.

The latest technology to enter this arena is Artificial Intelligence (AI), including a class of computational approaches involving large language models (LLMs) to make veracity judgments. Several empirical studies have investigated how well deceptive and truthful statements can be detected using different LLMs, and the results are mixed (e.g., Markowitz & Hancock, 2024; Markowitz & Levine, 2025; Miah et al., 2025). Such investigations have included several moderators for detection accuracy, including the idea of creating different AI personas to detect deception (Markowitz & Levine, 2025), varying the prompts to elicit a judgment (Markowitz & Hancock, 2024), and altering the evidence given to LLMs prior to their judgments (Miah et al., 2025). There is little consensus on the degree to which LLMs can effectively detect deception, and therefore, we seek to build on such research in several meaningful ways. Chief among them is that, unlike in prior work, we used seven theory-based Retrieval-Augmented Generation (RAG) frameworks to guide deception judgments by the LLMs. RAGs offer techniques to look for an answer within source materials prior to making a deception judgment. Therefore, we attempted to understand how lie-truth judgments are made using theory-based RAGs relative to baseline (non-RAG) models.

This work is timely and important because theory-based deception detection studies with AI are rare, and those that test theoretical principles tend to examine one theory in isolation. Our evaluation of seven leading deception theories helped us to understand those that might inform (or misinform) deception judgments when AI act as judges. To the best of our knowledge, the present work is one of the first to take a multi-theoretical approach to deception detection with AI. We also examined how well RAG-based models compare to baseline (non-RAG) models for the same datasets. This is an important technical interest because it is unclear if such extensive computation makes a difference in the accuracy or reliability of deception judgments. Finally, we also explore several moderating factors on deception judgments, including content effects (e.g., the idea that different deception datasets might lead to different veracity judgments and results) and model effects (e.g., the idea that different LLMs making such judgments might lead to different results). Some studies indicate content- and model-effects are present in similar evaluations (e.g., Markowitz & Hancock, 2024; Markowitz & Levine, 2025; Miah et al., 2025), but the results are far from conclusive and more evidence is needed.

Before we proceed, we believe it is important to be transparent about our aims with the current work. There is an inherent tension when a manuscript evaluates theories along the same dimensions (e.g., detection accuracy, truth-bias). There may be the appearance of a competition, where the seven theories are pitted against each other to find a "winner" (e.g., the theory that predicts veracity most accurately or in a less biased manner). We performed several steps to avoid this. First, the source materials for the RAGs were provided by each theorist where possible. Second, while the second author of this paper is also the author of a theory being tested, this theorist was not involved in the creation of the RAGs, nor the calculation or assessment of the results. Third, both authors of this paper collectively pride themselves on scientific objectivity. In other words, we "follow the data," regardless of whether the results are in favor of a theory they support or not. We believe this transparency is necessary to avoid the appearance of "tipping the scale" in one theory's favor over other theories. Any reliable data point on how to detect deception is a useful data point, regardless of the theory that it supports.

**Seven Theoretical Perspectives Guiding Deception and Deception Detection**

We begin by introducing the seven theoretical perspectives represented in our current empirical effort. They are described in chronological order, from the oldest theoretical perspective (e.g., Ekman's leakage approach) to the most recent theoretical perspective (i.e., truth-default theory), with brief commentary on how they should perform on deception detection tasks like those in our paper.

Historically, the first theory of deception detection was Ekman's leakage approach (Ekman & Friesen, 1969), although we characterize a more recent version (Ekman, 2001). In high-stakes situations, felt emotions are leaked through micro facial expressions and behavioral indicators of cognitive effort. Deception is indicated by discrepancies between verbal content and expressive behavior. Because the data in our paper is text-based, the leakage approach is handicapped by media affordances (e.g., facial expressions unavailable). Nevertheless, the extent to which leakage might be verbally signaled, Ekman's approach should perform better on higher-stakes truths and lies (e.g., court transcripts, cheating tapes). A survey of deception researchers reports leakage is currently held in low regard for its diagnosticity (Luke et al., 2025).

The four-factor theory (FFT; Zuckerman et al., 1981) expanded on Ekman's theorizing to specify four mechanisms producing deception cues. It suggests that lying is signaled by behaviors associated with arousal, emotions, cognitive effort, and behavioral control. Like leakage, detection based on the four-factor theory should perform poorly in text-based lie detection. Following the publication of DePaulo et al. (2003), which suggests most deception cues are faint and unreliable, FFT has largely been abandoned.

Information Manipulation Theory (IMT; McCornack, 1992), and later IMT2 (McCornack et al., 2014), was the first theory of deception to eschew nonverbal behavior and tackle verbal antecedents of deception. Information manipulation theory focuses on explaining why messages are deceptive (covert violations of Grice's maxims), the dimensions along which deceptive messages vary (quality, quantity, relevance, and manner) (Grice, 1975), and the theory attends to deceptive speech production rather than deception detection. Nevertheless, it predicts that omission is the most common means of deception. It is plausible that LLMs can detect or infer what is omitted, evaded, or obscured through language.

Interpersonal Deception Theory (IDT; Buller & Burgoon, 1996) is a further expansion of ideas from the four-factor theory, but adds a verbal component of information management strategies, and highlights dynamic and interactive aspects of the deception detection process. Like prior theories, its performance should be impaired in text-based analysis, but to a lesser extent. It should perform better when there is interaction between an interviewer and interviewee, and when answers are longer.

The verifiability approach (Nahari et al., 2014b, 2014a) is a fully verbal approach well suited to text-based lie detection. This approach suggests honest people want to be vindicated and will therefore provide verifiable details about an event or account, while deceivers want to be believed (but not detected) and therefore provide unverifiable details. Because it matters if the details are, in principle, *verifiable* rather than *verified*, the verifiability approach lends itself well to the current detection tasks.

Another approach lending itself to the tasks at hand is details and complications (Vrij et al., 2018, 2021). Both details and complications are verbal cues to deception. Compared to deceptive communication, honest communication is specified to be more detailed and contain more complications (characteristics that add to complexity). Details are considered the most commonly accepted and well-supported deception cue (Luke et al., 2025).

Finally, Truth-Default Theory (TDT; Levine, 2014, 2020) holds that honesty and passive belief are the default modes of human communication. Because most communication is honest and people tend to believe, detection efforts using this theory should perform better on truths than lies. The two best ways to detect deception (evidence and confessions) are precluded in current detection tasks. TDT is further hampered by the high rate of lies in current task relative to truths, relative to everyday communication. Having outlined these core theoretical ideas, we now turn to deception detection with AI.

**Artificial Intelligence and Deception Detection**

Despite a wealth of studies that have evaluated deception detection abilities in humans (see Bond & DePaulo, 2006; Levine & Serota, 2025), a minority of studies have assessed deception detection abilities in AI. Note that, before we review research that has used AI to detect deception, we distinguish between AI-based deception detection and NLP-based deception detection, as the latter has received significant treatment using various computational approaches (Constâncio et al., 2023). For example, a common interest in NLP-based deception detection is to build classifiers from language features (e.g., style words, content words, linguistic structure) to determine how well lies can be distinguished from truths. NLP-based deception detection can sometimes outperform human judgments (Kleinberg & Verschuere, 2020), but rarely to the degree that such models encourage scholars to abandon the human side of deception detection (Hauch et al., 2015; Lai & Tan, 2019).

The introduction of LLMs to academic spaces has facilitated new studies to examine the degree to which AI can detect deception relative to humans (King & Neal, 2024). There are several noteworthy papers in this space. Recent work by Miah et al. (2025) evaluated the data from real-life trial interviews (Pérez-Rosas et al., 2015), hotel reviews (Ott et al., 2011), and opinions about friends (Lloyd et al., 2019) to assess how well LLMs like *gpt-4o* could discriminate between lies and truths. The authors observed that deception detection accuracy was slightly above chance for data regarding hotel reviews and opinions about friends, but nearly 80% accuracy was achieved for real-life trials. Zero-shot results, in which a model receives "a task description prompt and input data without labeled examples" (Miah et al., 2025, p. 31016), revealed worse accuracy than few-shot results, which included labeled examples of lies and truths. Other research has used a range of LLMs to evaluate detection accuracy and other critical deception detection metrics like truth-bias (e.g., the probability of judging a message as true independent of actual message veracity), truth accuracy (e.g., among truthful trials, accuracy for truths), and deception accuracy (e.g., among deceptive trials, accuracy for deceptions). For example, Markowitz and Hancock (2024) observed that humans and AI (i.e., *gpt-3.5*, *gpt-4*, and *Bard*, which is currently *Gemini*) achieved deception detection accuracies that were remarkably consistent with each other, and AI were almost perfectly truth-biased when presented with a neutral prompt and a prompt trying to elicit suspicion. These LLMs were also substantially truth-biased when given the genuine base-rate of lies-to-truths in the sample. Finally, a more recent paper using audiovisual data of cheating interviews and opinions about friends observed critical content effects for LLM-based deception detection. Markowitz and Levine (2025) observed that, consistent with prior work (Markowitz & Hancock, 2024), *gemini-1.5-flash* was truth-biased when evaluating opinions-about-friends data, but this model was substantially lie-biased when evaluating videos of cheating interrogations. AI personas (e.g., an FBI agent vs. an undergraduate) did not lead to meaningful differences in detection accuracy nor response bias. Content effects and model effects, as the prior collective evidence suggests, are important moderators of deception judgments.

**The Current Paper**

There is a small collection of studies that have examined how well AI can detect deception, and no consensus exists about their utility or reliability. More research is required to examine how well such models can discriminate between lies and truths, and it is also critical to examine new LLM capabilities as they arise for deception detection. For example, in this work, we use RAG frameworks to guide deception judgments drawing on seven dominant theories in deception research. This is a critical advancement to the field because, thus far, we are unaware of studies that have examined the theoretical principles of deception theory in detection studies (albeit see Markowitz & Hancock, 2024; Markowitz & Levine, 2025 for evidence with truth-default theory). While the computation underlying a RAG framework is complex, its logic is simple, and a metaphor of a "cheat sheet" is helpful to demonstrate its capabilities. Imagine two students, with relatively equivalent abilities, who did not study for an upcoming exam. One student takes the exam based on intuition and their existing knowledge on a topic, while the second student can use a "cheat sheet" to look up answers from source documents. All else being equal, we would expect the student who used the source documents to perform better than the student who did not. However, this contention has been untested in deception detection research when using theory to guide RAGs.

Our current undertaking addresses this open question directly by developing seven RAGs based on deception theories and theoretical perspectives (presented in alphabetical order): (1) details and complications (Vrij et al., 2018, 2021), (2) Ekman's leakage approach (Ekman, 2001), (3) the Four-factor theory (Zuckerman et al., 1981), (4) Information Manipulation Theory (McCornack, 1992; McCornack et al., 2014), (5) Interpersonal Deception Theory (Buller & Burgoon, 1996), (6) Truth-Default Theory (Levine, 2014, 2020), and (7) the verifiability approach (Nahari et al., 2014b, 2014a). We use existing frameworks on how to successfully develop RAGs for social scientific investigations to build our seven theory-guided deception detection models (Bailenson et al., 2026), and examined how the models compared with and without RAG-based judgments. These interests led to two interrelated research questions:

$RQ_{1a}$: How do different theoretical models compare in deception judgments?

$RQ_{1b}$: How do RAG and baseline run-types compare across different theoretical models for deception judgments?

Recall, based on prior work (Markowitz & Hancock, 2024; Markowitz & Levine, 2025; Miah et al., 2025), we also anticipated two moderators for deception judgments using AI: (1) content effects and (2) model effects. First, to address content effects, we selected datasets that varied the deceptions people reported on and the level of interactivity of such deceptions. For example, our five datasets included responses from participants who denied cheating during an interview setting (Levine, 2007), those who reported falsely or truthfully about their opinions on friends (Lloyd et al., 2019), defendants in a real-life court trial (Pérez-Rosas et al., 2015), and false and truthful hotel reviews (Ott et al., 2011). We also included complete interview transcripts that reflected an interaction (e.g., the interviewer and interviewee), not just one side of an interaction (via Levine et al., 2014), given that certain theories argue that deception is a dynamic and interactive process between communicators (Buller & Burgoon, 1996; Burgoon, 2015). Second, we explored how different LLMs in the judgment stage of this process produced differential results. Different models have different reasoning abilities (Hagendorff et al., 2023; Moreira, 2026; Parmar et al., 2024), suggesting it is also important to explore how various LLMs perform on such judgment and decision-making tasks like deception detection.

> $RQ_2$: How do moderators like deception content and the LLM processing such texts relate to deception judgments?

## Method

One of our major interests was comparing the detection judgments of theory-specific RAGs or baseline models (no RAGs). To systematically achieve this goal, we drew on the work of Bailenson and colleagues (2026), who created a framework outlining the RAG-development process in terms of the Seven C's: *collecting* source materials, *cleaning* the source materials, *classifying* source materials (if they differ widely), *chunking* or disaggregating data into pieces, *creating* embeddings by transforming chunks into numerical vectors, *correlating* embeddings into a searchable database, and *connecting* to an LLM via an API to test deception judgments. Backend RAG development was completed with the assistance of Claude Code.

### Collecting

Our focus on using theory to guide the creation and testing of RAG-based deception detection resulted in seven candidate theories or theoretical positions: (1) details and complications (Vrij et al., 2018, 2021), (2) Ekman's leakage approach (Ekman, 2001), (3) the four-factor theory (Zuckerman et al., 1981), (4) Information Manipulation Theory (McCornack, 1992; McCornack et al., 2014), (5) Interpersonal Deception Theory (Buller & Burgoon, 1996), (6) Truth-Default Theory (Levine, 2014, 2020), and (7) the verifiability approach (Nahari et al., 2014b, 2014a). We selected these theories because they have been the dominant perspectives guiding much of deception detection over the past several decades, and they have also been the theories to receive some of the most extensive empirical testing in the field via meta-analyses, systematic reviews, and primary studies.

We began by soliciting the expertise of each theory developer or a scholar most closely aligned with the theory's contentions. These authors were contacted via email, explaining the purpose and direction of the project, and asked to provide up to 10 of the theory's most important works to be used as sources for the RAGs. Two of the authors were unable to be reached, and one declined to participate. Therefore, in such cases, the current authorship team selected the most important works from these theories. We obtained 50 total sources for all the RAGs (see Supplementary Table S1).

**Cleaning and Classifying**

Texts from each source were extracted and then subjected to several automated data cleaning processes. Two categories of content were targeted for removal: (a) tables, and (b) reference lists. Tables were excluded because tabular data (e.g., statistical results, coding schemes) are not well-suited to semantic retrieval. Table notes were retained. Reference lists were excluded because they contribute citation metadata rather than theoretical content.

Appendices and supplementary materials sections were retained. The cleaning pipeline was subjected to multiple independent verification passes, including (a) word-level differences between raw and cleaned texts to flag any removals that could not be attributable to a table or reference, (b) a page-level completeness check comparing the distinctive vocabulary of each source page against the final cleaned corpus, (c) a scan of the final several hundred words of each document to catch remaining reference material, and (d) manual visual spot-checks, in which rendered page images were directly compared against the corresponding cleaned text. Therefore, to the best of our ability, the data were cleaned and prepared for RAG creation. Since all source materials were academic works, the *classifying* step was not required and thus skipped.

**Chunking, Creating Embeddings, and Correlating Index**

Cleaned documents were segmented into retrievable units using a two-stage process. Each document was first split at heading boundaries to produce section-level units that preserved the source's structure and carried section-title metadata. For example, the heading-based pass isolated Vrij's (2019) subsection "3.1 | Within-subjects measurements" as a single section-level unit, tagged with that section title as metadata. The semantic parser then partitioned this unit into eight retrievable chunks ranging from 170 to 2,466 characters, with boundaries related to topic transitions (e.g., separating a chunk that discussed (a) cutoff-score thresholds for verifiable-detail ratios from (b) the subsequent chunk introducing the complications typology) rather than at fixed token or character intervals. Each resulting chunk retained its sections' title as retrievable metadata, preserving the original document structure even after fine-grained parsing.

Chunks were embedded using OpenAI's *text-embedding-3-large* model and maintained in a theory-specific ChromaDB vector store. One collection existed per theory/RAG and across the seven theories, the final corpus contained a total of 5,808 chunks.

Each RAG searched its theory-specific source documents using two different but simultaneous approaches and then combined the results. The first was a keyword search, searching for chunks that shared the same key terms as the judged statement (the "BM25" method). The second approach is a meaning-based search, which compares the meaning of the statement being judged to the meaning of each chunk. Each search method produced its own ranked list of the most relevant chunks. Those two lists were then merged into a single ranked list using a method called Reciprocal Rank Fusion (Cormack et al., 2009), which rewards chunks that scored well on either list, especially both. The top six fused chunks were passed to the LLM as retrieved context for each veracity judgment. The number of chunks is researcher-determined, and there is mixed evidence regarding the optimal number of chunks for different tasks (Bailenson et al., 2026; Cuconasu et al., 2024; Liu et al., 2023; Mazuryk et al., 2026).

**Connecting to an LLM**

To evaluate deception judgments, we developed a Graphical User Interface that allowed us to select an LLM to process the queries via APIs. We evaluated the performance of four LLMs in this paper: *gpt-4o* (OpenAI), *claude-sonnet-4-6* (Anthropic), *ollama/llama3* (Meta), and *deepseek-v4-flash* (DeepSeek).

The queries in this study consisted of two prompts, one that is system-level and one that is task-level. The system-level prompt identified each RAG/theory's core principles (see online supplement for verbatim descriptions). The second, task-level prompt presented the statement to be judged and instructed the model to commit to a binary veracity judgment (i.e., truthful or deceptive); see online supplement for verbatim instructions. When using RAG models, the task-level prompt was preceded by a block of retrieved passages returned by the retrieval process, and the output format included an additional field requiring the model to cite which retrieved passage(s) informed its judgment. When baseline models were applied, the task-level prompt was otherwise identical but omitted both the retrieved-passage block and the evidence-citation field, isolating the effect of retrieved context on deception judgments.

**Deception Datasets**

To examine how well RAG and non-RAG models compare for theory-based deception detection, we collected five diverse and published samples of text. The samples included truthful and deceptive hotel reviews (Ott et al., 2011), opinions about friends (Lloyd et al., 2019), transcripts from public court trials (Pérez-Rosas et al., 2015), and two studies containing interview transcripts where students denied cheating during a trivia game (Levine, 2007), one of which contained interactive sender and receiver data (Levine et al., 2014). We aimed to evaluate at most 200 cases from each dataset with an even split of ground-truth deceptive and truthful trials (we used all available cases if < 200). If there were more than 200 cases in a dataset, a random selection of 100 truthful and 100 deceptive statements from that dataset was selected (final $N = 700$). Our total number of judgments was 39,200 (700 trials × 7 theories × 4 LLMs × 2 run-types). Descriptive information about each dataset is offered in Supplementary Table S2.

**Analytic Plan**

We analyzed these data in several ways. First, we evaluated overall trends in accuracy, truth-bias, truth accuracy, and lie accuracy aggregated across run-types (e.g., RAG vs. baseline), datasets, and LLMs. Second, we disaggregated the results by (a) theory, (b) run-type, (c) theory and run-type, (d) theory, run-type, and dataset. Results disaggregated by LLMs are briefly mentioned below and are in the Supplementary Table S3 due to space considerations.

## Results

Consistent with prior work (Markowitz & Hancock, 2024) and human performance

(Bond & DePaulo, 2006), overall detection accuracy was 54.6% (*SD* = 49.8%) with a truth-bias of 58.3% (*SD* = 49.3%). Accuracy results aggregated by theory ranged from 50.9% (IDT) to 58.8% (details and complications), and the omnibus test was statistically significant [$F(6, 39193) = 18.67$, $p < .001$, $\eta^2 = .003$; see Table 1]. The details and complications accuracy results were significantly higher than other theoretical models (*p*s ≤ .002) except for TDT ($p = .961$; Tukey-corrected). Regarding truth-bias, results ranged from 32.2% (the verifiability approach) to 88.1% (TDT), and the omnibus test was statistically significant [$F(6, 39193) = 774.8$, $p < .001$, $\eta^2 = .11$]. In other words, judgments ranged from being highly lie-biased (the verifiability approach) to highly truth-biased (TDT). All comparisons for truth-bias were statistically significant except for the FFT-IDT comparison ($p = .993$).[1]

The results disaggregated by run-type (RAG vs. baseline) suggested accuracies were not statistically different from each other ($t = 0.22$, $p = .823$, Cohen's $d = .002$; see Table 2). Baseline models were more truth-biased than RAG models, on average, but the effect size is small ($t = 5.44$, $p < .001$, Cohen's $d = 0.06$).

**Content Effects**

To what degree are the accuracy and truth bias results moderated by the content of the judgments? We performed Theory × Dataset interaction effects, and the omnibus tests for accuracy [$F(24, 39165) = 34.01$, $p < .001$, $\eta^2_p = 0.02$] and truth bias [$F(24, 39165) = 121.20$, $p < .001$, $\eta^2_p = 0.07$] were both statistically significant. The evidence in Figure 1 suggests detection accuracy and truth-bias are substantially moderated by the content of the deceptions. For example, accuracy using the cheating interviews (Levine, 2007) varied by over 40 percentage points across theories, whereas performance for the Miami friend opinions data (Lloyd et al., 2019) and real-life trial data (Pérez-Rosas et al., 2015) remained comparatively stable. Regarding truth-bias, the TDT-based models generally increased the tendency to classify messages as truthful independent of actual message veracity.

---

[1] On average, the *claude-sonnet-4-6* model was more accurate ($p < .001$) and more truth-biased ($p < .001$) than *gpt-4o*, *llama3,* and *deepseek-v4-flash*. There were no significant differences in accuracy nor truth-bias between *gpt-4o* and *llama3* (*p*s > .223).

A three-way interaction of Theory × Dataset × Run-type was not statistically significant for accuracy, [$F(24, 39130) = 1.14$, $p = .292$, $\eta^2_p = 6.97\text{e-}04$], but it was statistically significant for truth-bias, [$F(24, 39130) = 2.16$, $p < .001$, $\eta^2_p = 1.33\text{e-}03$]. Specifically, as Figure 2 displays, TDT produced comparatively high truth bias scores across nearly every dataset (RAG-to-baseline comparisons all *p*s > .351), whereas the verifiability approach became more lie-biased under RAG in four of five datasets, with only the interactive cheating interviews moving in the opposite direction toward truth-bias. See Table 3 for additional details.

**Exploratory Reasoning Results**

While soliciting deception judgments, we also instructed the LLMs to provide a rationale explaining why such judgments were made. These reasonings were extracted and submitted to a natural language processing analysis to understand how much the thinking styles of the LLMs varied across theories. We applied the text analysis program, Linguistic Inquiry and Word Count (LIWC: Pennebaker et al., 2022), and used two established measures of cognition that prior work suggests help to understand how individuals are engaged in meaning-making processes.

The first dimension is analytic thinking (Pennebaker et al., 2014), which measures the degree to which a communicative style is formal and hierarchically structured. Analytic thinking is based on a collection of eight style word categories, which factor-load onto a single dimension: high rates of articles and prepositions, but low rates of impersonal pronouns, personal pronouns, auxiliary verbs, adverbs, conjunctions, and negations. Prior work suggests high rates of analytic thinking are associated with one's interest and excitement around deep thinking (e.g., need for cognition; Markowitz, 2023) and has been connected to a range of social and psychological dynamics like cognitive load (Seraj et al., 2021) and distress (Markowitz, 2022). The analytic thinking measure is a proxy for Kahneman's (2011) System 2, which describes deliberate, rational, and slow thinking, and is measured on a standardized scale from 0 (low analytic thinking) to 100 (high analytic thinking). Scores near 50 are "middle of the road."

The second dimension of interest is cognitive processing terms, which measures the degree to which an individual is "working through" an issue and trying to make meaning from it (Boyd et al., 2020; Pennebaker et al., 2003). Words like *because*, *understand*, and *know* indicate how much a person has resolved an issue in their mind and how much they have not. For example, people reporting on romantic distress tend to feature a spike in their use of cognitive processing terms at the time they were broken up with relative to a baseline period (Seraj et al., 2021). Cognitive processing terms are counted as a percentage of the total word count per text. Together, analytic thinking and cognitive processing terms are two dimensions that approximate how people are thinking and working through a problem or decision, and we apply them to examine how LLMs thought through their deception judgments.

Supplementary Tables S4 and S5 report the results aggregated by theory. The omnibus tests for analytic thinking [$F(6, 39188) = 362.90$, $p < .001$, $\eta^2 = 0.05$] and cognitive processing [$F(6, 39193) = 598.50$, $p < .001$, $\eta^2 = 0.08$] were statistically significant.[2] The most formal and analytic rationales concerned IDT, IMT, and TDT, while the most informal and narrative-like rationales concerned details and complications plus the verifiability approach. By contrast, these theories also contained some of the lowest and highest rates of cognitive processing terms, respectively. Combined with the detection accuracy, it is striking to observe that the theories with the highest detection accuracy (e.g., details and complications, the verifiability approach) were associated with rationales containing the greatest number of cognitive effort indicators. Perhaps these insights suggest LLM effort, as represented in language, is a characteristic that is linked with deception judgments. Fully disaggregated results by theory, dataset, run-type, and model are in Supplementary Tables S6 and S7 due to space considerations.

---

[2] Five cases did not contain rationales and therefore, they were dropped from related analyses.

## Discussion

In the current paper, we developed RAG frameworks based on deception theories to identify how well such AI models compared to baseline (non-RAG) models. On average, RAG-based and non-RAG-based models were not statistically different in detection accuracy, yet RAGs were slightly less truth-biased than non-RAGs. There was substantial heterogeneity in accuracy and truth-bias across theoretical perspectives as well. Theories that obtained the highest accuracy, on average, were details and complications, the verifiability approach and TDT, yet TDT was also the most truth-biased model. Moderators of accuracy and truth-bias, including content effects and model effects, shifted the results in meaningful ways as well. For example, depending on the theory-based RAG under examination, accuracy for cheating interview transcripts ranged between 30% and 73%. The Anthropic-based LLM (i.e., *claude-sonnet-4-6*) tended to also be the most accurate and most truth-biased model versus other LLMs.

In many ways, the results of the present paper are unsurprising. Primary studies have demonstrated the unreliability and contingent nature of deception judgments using out-of-the-box AI models (Markowitz & Hancock, 2024; Markowitz & Levine, 2025; Miah et al., 2025), yet we offer an important nuance to these results. What was unclear prior to the current investigation was the degree to which deception judgments were impacted by how judgments were made. Using RAGs, which referenced source documents prior to making a deception judgment, did not improve detection accuracy and only slightly reduced truth-bias (i.e., by approximately 3%, on average). In the long run, perhaps the truth-bias result is an important one. For example, over millions of trials and judgments, it might be useful to employ RAGs for deception detection and avoid such a response bias. A fact-checking AI model that assesses the veracity of millions of headlines or social media posts in real-time might benefit from a RAG in not presuming a message's honesty independent of its actual veracity. However, in the current ecosystem of AI models and how they are used, researchers and practitioners should be cautious about applying AI for deception judgments and expecting universal improvement in accuracy and response bias. Like other studies have suggested, deception detection with AI models — RAG-based or otherwise — is not ready for primetime in a forensic or applied sense.

Despite RAGs failing to make an appreciable difference in deception judgments relative to baseline models, there are several important contributions of this research. First, and most consequentially, our results highlight the importance of looking "under the hood" at deception detection results and separating discriminability from response bias. That is, the theory under examination did little to change how well models discriminated lies from truths, but it changed response bias. Accuracy across the seven theoretical perspectives spanned roughly eight percentage points, and the associated effect size was negligible ($\eta^2 = .003$), whereas truth-bias spanned fifty-six percentage points with an effect size nearly forty times larger ($\eta^2 = .11$). Put another way, providing an AI model with a theory of deception did not offer more diagnostic information about the message. Instead, it gave such AI models a decision rule about how to make judgments in the absence of diagnostic information, which varied from theory to theory.

Second, the current work provides an important empirical test of which theories showed discriminability and which did not in text-based deception detection. We argued that theories built on nonverbal and physiological indicators would struggle in a text-only task and that more verbal frameworks would benefit. The results are consistent with this assessment, where details and complications and the verifiability approach were among the most accurate and least truth-biased models (though the verifiability approach model was highly lie-biased). Ekman's leakage approach, the four-factor theory, IMT, and IDT, on the other hand, were among the models that struggled the most with detection accuracy. We believe at least two outcomes may have occurred, and they are not necessarily in tension with each other: (1) there are some theories that perform better in AI detection tasks than others, and (2) there are some detection tasks that facilitate success for certain theories better than others. Testing a theory whose core constructs may be unavailable in the data does not inherently take away from the theory's ability to be diagnostic. Perhaps a greater theory-to-data match at the theory level would be helpful for future studies. A related and more granular observation supports this conclusion. IDT, which argues that deception is dynamic and interactive, performed better on the interactive cheating interviews (43.3% baseline, 42.2% RAG) than on the non-interactive cheating interviews (32.4% baseline, 28.8% RAG), even though both remained below chance. Future studies should be thoughtful about theory-to-data (mis)matches when interpreting the results of their work.

A third contribution of this work is the articulation of *theory* as a potential experimental factor. By transforming seven deception theories into RAGs, theory becomes a variable that can be assigned, crossed with other factors, and evaluated at a scale that would be otherwise difficult to perform with humans. This inverts the usual relationship between theory and empirical evaluation. Rather than stating what a theory predicts and then testing its predictions, we can evaluate a theory and its core ideas across thousands of judgments to observe how it operates using AI models as judges or participants. This ability is not confined to deception, yet soliciting source materials from theory developers is an underappreciated component of the study's design, and one we would encourage others to repeat in similar investigations.

Finally, our exploratory natural language processing results point to one potential explanation for some of the effects reported in this paper: cognitive effort, and the degree to which the LLMs "worked through" the deception judgments. Other scholarship has evaluated AI rationales in deception detection for key themes (Markowitz & Levine, 2025), but our work presents some of the first evidence investigating the cognitive patterns of LLMs in deception judgments via language. Cognitive effort and cognitive approaches are a highly debated topic in deception (e.g., Levine et al., 2018; McCornack et al., 2014; Vrij et al., 2017), however, suggesting the contentions raised in the exploratory analyses deserve additional scrutiny.

**Limitations and Future Directions**

We used a limited number of datasets and LLMs in our study. Therefore, a natural next step would be to recruit more (theory-aligned) datasets and test the performance of more LLMs. Especially needed are audio-visual data for the theories prioritizing nonverbal indicators of deception. Second, the number of chunks was six in our RAG-based models, and there is no consensus as to the number of chunks that should be used in a task like ours. Future work should examine number of chunks (and chunk size) as another factor of the research design, though prior work suggest number of chunks did not make a meaningful difference in social scientific outcome like performance on an academic test (i.e., 3, 5, or 7 chunks; Bailenson et al., 2026).

Deception judgments were also made using text. Some work has found results that are consistent across modalities (e.g., in text-based and audiovisual evaluations of opinions about friends, detection accuracy tends to be low and AI models are truth-biased, on average;

Markowitz & Hancock, 2024; Markowitz & Levine, 2025). However, future work should also use stimuli that prize nonverbal or dynamic aspects of deception. Finally, the approach taken in the current work used zero-shot results and this was purposeful, as deception detection tasks with humans rarely give examples to judges. Future studies should examine how zero-shot versus few-shot results compare when evaluating RAGs vs. baseline models.

## Acknowledgments

The authors thank [REDACTED] for their assistance in providing source materials for their respective theory-based RAGs.

**Table 1**

*Results Aggregated by Theory*

| | Accuracy | | | Truth bias | | | Truth accuracy | | | Lie accuracy | | |
|---|---|---|---|---|---|---|---|---|---|---|---|---|
| Theory | *M* | *SD* | 95% CI | *M* | *SD* | 95% CI | *M* | *SD* | 95% CI | *M* | *SD* | 95% CI |
| D&C | 58.8% | 49.2% | [57.5, 60.1] | 46.9% | 49.9% | [45.6, 48.3] | 54.8% | 49.8% | [53.1, 56.5] | 64.6% | 47.8% | [62.6, 66.6] |
| Ekman | 53.2% | 49.9% | [51.9, 54.6] | 55.7% | 49.7% | [54.4, 57.0] | 57.6% | 49.4% | [55.9, 59.2] | 46.9% | 49.9% | [44.9, 49.0] |
| FFT | 53.2% | 49.9% | [51.9, 54.6] | 58.3% | 49.3% | [57.0, 59.6] | 59.7% | 49.1% | [58.1, 61.4] | 43.8% | 49.6% | [41.7, 45.8] |
| IDT | 50.9% | 50.0% | [49.6, 52.2] | 58.9% | 49.2% | [57.6, 60.2] | 58.3% | 49.3% | [56.6, 60.0] | 40.1% | 49.0% | [38.1, 42.2] |
| IMT | 52.7% | 49.9% | [51.4, 54.0] | 68.2% | 46.6% | [66.9, 69.4] | 67.6% | 46.8% | [66.0, 69.1] | 31.0% | 46.3% | [29.1, 32.9] |
| TDT | 57.9% | 49.4% | [56.6, 59.2] | 88.1% | 32.4% | [87.2, 88.9] | 88.7% | 31.7% | [87.6, 89.7] | 12.8% | 33.4% | [11.5, 14.2] |
| VA | 55.2% | 49.7% | [53.9, 56.5] | 32.2% | 46.7% | [31.0, 33.4] | 39.4% | 48.9% | [37.7, 41.0] | 78.3% | 41.2% | [76.6, 80.0] |

*Note*. D&C = Details and complications; FFT = Four-Factor Theory, IDT = Interpersonal Deception Theory, IMT = Information Manipulation Theory, TDT = Truth-Default Theory, VA = Verifiability Approach. 95% CIs are Wilson score intervals, which remain bounded by 0 and 100%. For such outcomes SD = $\sqrt{[p(1 - p)]}$ is a function of the mean, whereas 95% CIs reflect the number of judgments in each cell. $n$ = 5,600 per theory.

**Table 2**

*Aggregated and Disaggregated RAG and Baseline Results*

| | | Accuracy | | | Truth bias | | | Truth accuracy | | | Lie accuracy | | |
|---|---|---|---|---|---|---|---|---|---|---|---|---|---|
| | Run-type | *M* | *SD* | 95% CI | *M* | *SD* | 95% CI | *M* | *SD* | 95% CI | *M* | *SD* | 95% CI |
| | Baseline | 54.6% | 49.8% | [53.9, 55.3] | 59.7% | 49.1% | [59.0, 60.4] | 62.1% | 48.5% | [61.2, 62.9] | 43.8% | 49.6% | [42.7, 44.9] |
| | RAG | 54.5% | 49.8% | [53.8, 55.2] | 57.0% | 49.5% | [56.3, 57.7] | 59.7% | 49.1% | [58.8, 60.6] | 47.0% | 49.9% | [45.9, 48.1] |
| | | Accuracy | | | Truth bias | | | Truth accuracy | | | Lie accuracy | | |
| Theory | Run-type | *M* | *SD* | 95% CI | *M* | *SD* | 95% CI | *M* | *SD* | 95% CI | *M* | *SD* | 95% CI |
| D&C | Baseline | 59.4% | 49.1% | [57.6, 61.2] | 50.4% | 50.0% | [48.6, 52.3] | 58.3% | 49.3% | [55.9, 60.6] | 61.1% | 48.8% | [58.2, 63.9] |
| D&C | RAG | 58.2% | 49.3% | [56.3, 60.0] | 43.5% | 49.6% | [41.6, 45.3] | 51.4% | 50.0% | [49.0, 53.8] | 68.1% | 46.6% | [65.4, 70.8] |
| Ekman | Baseline | 52.9% | 49.9% | [51.0, 54.7] | 56.3% | 49.6% | [54.5, 58.1] | 57.8% | 49.4% | [55.4, 60.1] | 45.8% | 49.8% | [42.9, 48.7] |
| Ekman | RAG | 53.6% | 49.9% | [51.8, 55.4] | 55.1% | 49.7% | [53.3, 57.0] | 57.4% | 49.5% | [55.0, 59.7] | 48.1% | 50.0% | [45.2, 51.0] |
| FFT | Baseline | 52.9% | 49.9% | [51.0, 54.7] | 60.9% | 48.8% | [59.0, 62.6] | 61.5% | 48.7% | [59.2, 63.8] | 40.1% | 49.0% | [37.3, 43.0] |
| FFT | RAG | 53.6% | 49.9% | [51.8, 55.5] | 55.8% | 49.7% | [53.9, 57.6] | 57.9% | 49.4% | [55.5, 60.3] | 47.4% | 50.0% | [44.5, 50.3] |
| IDT | Baseline | 51.0% | 50.0% | [49.2, 52.9] | 60.2% | 49.0% | [58.4, 62.0] | 59.4% | 49.1% | [57.1, 61.8] | 38.7% | 48.7% | [35.9, 41.6] |
| IDT | RAG | 50.8% | 50.0% | [49.0, 52.7] | 57.7% | 49.4% | [55.8, 59.5] | 57.2% | 49.5% | [54.8, 59.5] | 41.5% | 49.3% | [38.7, 44.4] |
| IMT | Baseline | 51.9% | 50.0% | [50.0, 53.7] | 66.7% | 47.1% | [64.9, 68.4] | 65.6% | 47.5% | [63.3, 67.9] | 31.8% | 46.6% | [29.1, 34.5] |
| IMT | RAG | 53.6% | 49.9% | [51.7, 55.4] | 69.6% | 46.0% | [67.9, 71.3] | 69.5% | 46.0% | [67.3, 71.7] | 30.2% | 45.9% | [27.6, 32.9] |
| TDT | Baseline | 57.7% | 49.4% | [55.9, 59.5] | 88.2% | 32.2% | [87.0, 89.4] | 88.6% | 31.7% | [87.0, 90.1] | 12.4% | 33.0% | [10.6, 14.5] |
| TDT | RAG | 58.1% | 49.4% | [56.2, 59.9] | 88.0% | 32.5% | [86.7, 89.2] | 88.8% | 31.6% | [87.2, 90.2] | 13.1% | 33.8% | [11.3, 15.2] |
| VA | Baseline | 56.6% | 49.6% | [54.8, 58.4] | 35.2% | 47.8% | [33.4, 37.0] | 43.1% | 49.5% | [40.7, 45.5] | 76.4% | 42.5% | [73.9, 78.8] |
| VA | RAG | 53.8% | 49.9% | [51.9, 55.6] | 29.2% | 45.5% | [27.5, 30.9] | 35.6% | 47.9% | [33.4, 38.0] | 80.3% | 39.8% | [77.9, 82.5] |

*Note*. D&C = Details and complications; FFT = Four-Factor Theory, IDT = Interpersonal Deception Theory, IMT = Information Manipulation Theory, TDT = Truth-Default Theory, VA = Verifiability Approach. 95% CIs are Wilson score intervals.

**Table 3**

*Results Disaggregated by RAG, Run, and Dataset*

| Theory | Run-type | Dataset | Accuracy | | | Truth bias | | | Truth accuracy | | | Lie accuracy | | | | | |
|---|---|---|---|---|---|---|---|---|---|---|---|---|---|---|---|---|---|
| | | | *M* | *SD* | 95% CI | *M* | *SD* | 95% CI | *M* | *SD* | 95% CI | *M* | *SD* | 95% CI | *n* | *Truthful n* | *Deceptive n* |
| D&C | Baseline | Hotel reviews | 70.6% | 45.6% | [67.4, 73.7] | 50.4% | 50.0% | [46.9, 53.8] | 71.0% | 45.4% | [66.4, 75.2] | 70.2% | 45.8% | [65.6, 74.5] | 800 | 400 | 400 |
| D&C | Baseline | Cheating 1 | 51.1% | 50.0% | [46.5, 55.7] | 51.6% | 50.0% | [46.9, 56.2] | 51.6% | 50.0% | [46.6, 56.6] | 48.6% | 50.3% | [37.4, 59.9] | 448 | 376 | 72 |
| D&C | Baseline | Cheating 2 | 69.8% | 46.0% | [64.0, 75.0] | 69.8% | 46.0% | [64.0, 75.0] | 71.4% | 45.3% | [65.5, 76.6] | 50.0% | 51.3% | [29.9, 70.1] | 268 | 248 | 20 |
| D&C | Baseline | Miami friend opinions | 51.7% | 50.0% | [48.3, 55.2] | 40.5% | 49.1% | [37.2, 43.9] | 42.2% | 49.5% | [37.5, 47.1] | 61.3% | 48.8% | [56.4, 65.9] | 800 | 400 | 400 |
| D&C | Baseline | Trial data | 55.6% | 49.7% | [51.1, 59.9] | 55.2% | 49.8% | [50.7, 59.5] | 60.8% | 48.9% | [54.5, 66.8] | 50.4% | 50.1% | [44.2, 56.6] | 484 | 240 | 244 |
| D&C | RAG | Hotel reviews | 65.6% | 47.5% | [62.3, 68.8] | 44.4% | 49.7% | [41.0, 47.8] | 60.0% | 49.1% | [55.1, 64.7] | 71.2% | 45.3% | [66.6, 75.5] | 800 | 400 | 400 |
| D&C | RAG | Cheating 1 | 48.0% | 50.0% | [43.4, 52.6] | 47.1% | 50.0% | [42.5, 51.7] | 47.1% | 50.0% | [42.1, 52.1] | 52.8% | 50.3% | [41.4, 63.9] | 448 | 376 | 72 |
| D&C | RAG | Cheating 2 | 69.0% | 46.3% | [63.3, 74.3] | 71.3% | 45.3% | [65.6, 76.4] | 71.8% | 45.1% | [65.9, 77.0] | 35.0% | 48.9% | [18.1, 56.7] | 268 | 248 | 20 |
| D&C | RAG | Miami friend opinions | 52.6% | 50.0% | [49.2, 56.1] | 30.1% | 45.9% | [27.0, 33.4] | 32.8% | 47.0% | [28.3, 37.5] | 72.5% | 44.7% | [67.9, 76.6] | 800 | 400 | 400 |
| D&C | RAG | Trial data | 58.5% | 49.3% | [54.0, 62.8] | 45.2% | 49.8% | [40.9, 49.7] | 53.8% | 50.0% | [47.4, 59.9] | 63.1% | 48.3% | [56.9, 68.9] | 484 | 240 | 244 |
| Ekman | Baseline | Hotel reviews | 63.0% | 48.3% | [59.6, 66.3] | 83.0% | 37.6% | [80.2, 85.4] | 96.0% | 19.6% | [93.6, 97.5] | 30.0% | 45.9% | [25.7, 34.7] | 800 | 400 | 400 |
| Ekman | Baseline | Cheating 1 | 33.3% | 47.2% | [29.1, 37.7] | 24.8% | 43.2% | [21.0, 29.0] | 25.0% | 43.4% | [20.9, 29.6] | 76.4% | 42.8% | [65.4, 84.7] | 448 | 376 | 72 |
| Ekman | Baseline | Cheating 2 | 36.2% | 48.1% | [30.7, 42.1] | 30.2% | 46.0% | [25.0, 36.0] | 31.9% | 46.7% | [26.4, 37.9] | 90.0% | 30.8% | [69.9, 97.2] | 268 | 248 | 20 |
| Ekman | Baseline | Miami friend opinions | 52.9% | 49.9% | [49.4, 56.3] | 66.1% | 47.4% | [62.8, 69.3] | 69.0% | 46.3% | [64.3, 73.3] | 36.8% | 48.3% | [32.2, 41.6] | 800 | 400 | 400 |
| Ekman | Baseline | Trial data | 63.6% | 48.2% | [59.3, 67.8] | 39.7% | 49.0% | [35.4, 44.1] | 53.3% | 50.0% | [47.0, 59.5] | 73.8% | 44.1% | [67.9, 78.9] | 484 | 240 | 244 |
| Ekman | RAG | Hotel reviews | 61.8% | 48.6% | [58.3, 65.1] | 82.2% | 38.2% | [79.5, 84.7] | 94.0% | 23.8% | [91.2, 95.9] | 29.5% | 45.7% | [25.2, 34.1] | 800 | 400 | 400 |
| Ekman | RAG | Cheating 1 | 35.9% | 48.0% | [31.6, 40.5] | 30.6% | 46.1% | [26.5, 35.0] | 30.1% | 45.9% | [25.6, 34.9] | 66.7% | 47.5% | [55.2, 76.5] | 448 | 376 | 72 |
| Ekman | RAG | Cheating 2 | 41.8% | 49.4% | [36.0, 47.8] | 35.8% | 48.0% | [30.3, 41.7] | 37.9% | 48.6% | [32.1, 44.1] | 90.0% | 30.8% | [69.9, 97.2] | 268 | 248 | 20 |
| Ekman | RAG | Miami friend opinions | 54.0% | 49.9% | [50.5, 57.4] | 60.2% | 49.0% | [56.8, 63.6] | 64.2% | 48.0% | [59.4, 68.8] | 43.8% | 49.7% | [39.0, 48.6] | 800 | 400 | 400 |
| Ekman | RAG | Trial data | 62.4% | 48.5% | [58.0, 66.6] | 35.3% | 47.8% | [31.2, 39.7] | 47.9% | 50.1% | [41.7, 54.2] | 76.6% | 42.4% | [70.9, 81.5] | 484 | 240 | 244 |
| FFT | Baseline | Hotel reviews | 60.5% | 48.9% | [57.1, 63.8] | 82.8% | 37.8% | [80.0, 85.2] | 93.2% | 25.1% | [90.4, 95.3] | 27.8% | 44.8% | [23.6, 32.3] | 800 | 400 | 400 |
| FFT | Baseline | Cheating 1 | 39.1% | 48.8% | [34.7, 43.7] | 35.5% | 47.9% | [31.2, 40.0] | 34.8% | 47.7% | [30.2, 39.8] | 61.1% | 49.1% | [49.6, 71.5] | 448 | 376 | 72 |
| FFT | Baseline | Cheating 2 | 36.2% | 48.1% | [30.7, 42.1] | 31.0% | 46.3% | [25.7, 36.7] | 32.3% | 46.8% | [26.7, 38.3] | 85.0% | 36.6% | [64.0, 94.8] | 268 | 248 | 20 |
| FFT | Baseline | Miami friend opinions | 52.2% | 50.0% | [48.8, 55.7] | 73.0% | 44.4% | [69.8, 76.0] | 75.2% | 43.2% | [70.8, 79.2] | 29.2% | 45.5% | [25.0, 33.9] | 800 | 400 | 400 |
| FFT | Baseline | Trial data | 63.2% | 48.3% | [58.8, 67.4] | 44.6% | 49.8% | [40.3, 49.1] | 57.9% | 49.5% | [51.6, 64.0] | 68.4% | 46.6% | [62.4, 73.9] | 484 | 240 | 244 |
| FFT | RAG | Hotel reviews | 63.6% | 48.1% | [60.2, 66.9] | 78.6% | 41.0% | [75.7, 81.3] | 92.2% | 26.8% | [89.2, 94.5] | 35.0% | 47.8% | [30.5, 39.8] | 800 | 400 | 400 |
| FFT | RAG | Cheating 1 | 37.3% | 48.4% | [32.9, 41.8] | 32.8% | 47.0% | [28.6, 37.3] | 32.2% | 46.8% | [27.7, 37.1] | 63.9% | 48.4% | [52.4, 74.0] | 448 | 376 | 72 |
| FFT | RAG | Cheating 2 | 42.5% | 49.5% | [36.8, 48.5] | 38.1% | 48.6% | [32.5, 44.0] | 39.5% | 49.0% | [33.6, 45.7] | 80.0% | 41.0% | [58.4, 91.9] | 268 | 248 | 20 |
| FFT | RAG | Miami friend opinions | 52.6% | 50.0% | [49.2, 56.1] | 62.1% | 48.5% | [58.7, 65.4] | 64.8% | 47.8% | [59.9, 69.3] | 40.5% | 49.2% | [35.8, 45.4] | 800 | 400 | 400 |
| FFT | RAG | Trial data | 60.1% | 49.0% | [55.7, 64.4] | 38.6% | 48.7% | [34.4, 43.0] | 48.8% | 50.1% | [42.5, 55.0] | 71.3% | 45.3% | [65.3, 76.6] | 484 | 240 | 244 |
| IDT | Baseline | Hotel reviews | 57.4% | 49.5% | [53.9, 60.8] | 91.6% | 27.7% | [89.5, 93.4] | 99.0% | 10.0% | [97.5, 99.6] | 15.8% | 36.5% | [12.5, 19.6] | 800 | 400 | 400 |
| IDT | Baseline | Cheating 1 | 32.4% | 46.8% | [28.2, 36.8] | 26.1% | 44.0% | [22.3, 30.4] | 25.3% | 43.5% | [21.1, 29.9] | 69.4% | 46.4% | [58.0, 78.9] | 448 | 376 | 72 |
| IDT | Baseline | Cheating 2 | 43.3% | 49.6% | [37.5, 49.3] | 39.6% | 49.0% | [33.9, 45.5] | 40.7% | 49.2% | [34.8, 46.9] | 75.0% | 44.4% | [53.1, 88.8] | 268 | 248 | 20 |
| IDT | Baseline | Miami friend opinions | 51.9% | 50.0% | [48.4, 55.3] | 69.4% | 46.1% | [66.1, 72.5] | 71.2% | 45.3% | [66.6, 75.5] | 32.5% | 46.9% | [28.1, 37.2] | 800 | 400 | 400 |

| | | | | | | | | | | | | | | | | |
|---|---|---|---|---|---|---|---|---|---|---|---|---|---|---|---|---|---|
| IDT | Baseline | Trial data | 60.7% | 48.9% | [56.3, 65.0] | 36.0% | 48.0% | [31.8, 40.3] | 46.7% | 50.0% | [40.5, 53.0] | 74.6% | 43.6% | [68.8, 79.6] | 484 | 240 | 244 |
| IDT | RAG | Hotel reviews | 58.2% | 49.3% | [54.8, 61.6] | 89.5% | 30.7% | [87.2, 91.4] | 97.8% | 14.8% | [95.8, 98.8] | 18.8% | 39.1% | [15.2, 22.9] | 800 | 400 | 400 |
| IDT | RAG | Cheating 1 | 28.8% | 45.3% | [24.8, 33.2] | 22.1% | 41.5% | [18.5, 26.2] | 20.7% | 40.6% | [17.0, 25.1] | 70.8% | 45.8% | [59.5, 80.1] | 448 | 376 | 72 |
| IDT | RAG | Cheating 2 | 42.2% | 49.5% | [36.4, 48.1] | 37.7% | 48.6% | [32.1, 43.6] | 39.1% | 48.9% | [33.2, 45.3] | 80.0% | 41.0% | [58.4, 91.9] | 268 | 248 | 20 |
| IDT | RAG | Miami friend opinions | 52.4% | 50.0% | [48.9, 55.8] | 68.1% | 46.6% | [64.8, 71.3] | 70.5% | 45.7% | [65.9, 74.8] | 34.2% | 47.5% | [29.8, 39.0] | 800 | 400 | 400 |
| IDT | RAG | Trial data | 61.2% | 48.8% | [56.7, 65.4] | 31.8% | 46.6% | [27.8, 36.1] | 42.9% | 49.6% | [36.8, 49.2] | 79.1% | 40.7% | [73.6, 83.7] | 484 | 240 | 244 |
| IMT | Baseline | Hotel reviews | 56.4% | 49.6% | [52.9, 59.8] | 92.4% | 26.6% | [90.3, 94.0] | 98.8% | 11.1% | [97.1, 99.5] | 14.0% | 34.7% | [10.9, 17.7] | 800 | 400 | 400 |
| IMT | Baseline | Cheating 1 | 38.4% | 48.7% | [34.0, 43.0] | 35.7% | 48.0% | [31.4, 40.3] | 34.6% | 47.6% | [29.9, 39.5] | 58.3% | 49.6% | [46.8, 69.0] | 448 | 376 | 72 |
| IMT | Baseline | Cheating 2 | 43.7% | 49.7% | [37.8, 49.6] | 40.7% | 49.2% | [35.0, 46.6] | 41.5% | 49.4% | [35.6, 47.7] | 70.0% | 47.0% | [48.1, 85.5] | 268 | 248 | 20 |
| IMT | Baseline | Miami friend opinions | 51.7% | 50.0% | [48.3, 55.2] | 76.8% | 42.3% | [73.7, 79.5] | 78.5% | 41.1% | [74.2, 82.2] | 25.0% | 43.4% | [21.0, 29.5] | 800 | 400 | 400 |
| IMT | Baseline | Trial data | 61.8% | 48.6% | [57.4, 66.0] | 50.6% | 50.0% | [46.2, 55.1] | 62.5% | 48.5% | [56.2, 68.4] | 61.1% | 48.9% | [54.8, 67.0] | 484 | 240 | 244 |
| IMT | RAG | Hotel reviews | 52.8% | 50.0% | [49.3, 56.2] | 96.2% | 19.0% | [94.7, 97.4] | 99.0% | 10.0% | [97.5, 99.6] | 6.5% | 24.7% | [4.5, 9.4] | 800 | 400 | 400 |
| IMT | RAG | Cheating 1 | 48.9% | 50.0% | [44.3, 53.5] | 52.0% | 50.0% | [47.4, 56.6] | 50.5% | 50.1% | [45.5, 55.6] | 40.3% | 49.4% | [29.7, 51.8] | 448 | 376 | 72 |
| IMT | RAG | Cheating 2 | 54.1% | 49.9% | [48.1, 60.0] | 51.9% | 50.1% | [45.9, 57.8] | 53.2% | 50.0% | [47.0, 59.3] | 65.0% | 48.9% | [43.3, 81.9] | 268 | 248 | 20 |
| IMT | RAG | Miami friend opinions | 51.6% | 50.0% | [48.2, 55.1] | 74.6% | 43.5% | [71.5, 77.5] | 76.2% | 42.6% | [71.8, 80.2] | 27.0% | 44.5% | [22.9, 31.6] | 800 | 400 | 400 |
| IMT | RAG | Trial data | 62.2% | 48.5% | [57.8, 66.4] | 43.6% | 49.6% | [39.2, 48.0] | 55.8% | 49.8% | [49.5, 62.0] | 68.4% | 46.6% | [62.4, 73.9] | 484 | 240 | 244 |
| TDT | Baseline | Hotel reviews | 51.6% | 50.0% | [48.2, 55.1] | 98.1% | 13.6% | [96.9, 98.9] | 99.8% | 5.0% | [98.6, 100.0] | 3.5% | 18.4% | [2.1, 5.8] | 800 | 400 | 400 |
| TDT | Baseline | Cheating 1 | 73.7% | 44.1% | [69.4, 77.5] | 84.4% | 36.3% | [80.7, 87.4] | 84.6% | 36.2% | [80.6, 87.9] | 16.7% | 37.5% | [9.8, 26.9] | 448 | 376 | 72 |
| TDT | Baseline | Cheating 2 | 68.7% | 46.5% | [62.9, 73.9] | 70.1% | 45.8% | [64.4, 75.3] | 71.0% | 45.5% | [65.0, 76.3] | 40.0% | 50.3% | [21.9, 61.3] | 268 | 248 | 20 |
| TDT | Baseline | Miami friend opinions | 50.6% | 50.0% | [47.2, 54.1] | 91.1% | 28.5% | [89.0, 92.9] | 91.8% | 27.5% | [88.6, 94.1] | 9.5% | 29.4% | [7.0, 12.8] | 800 | 400 | 400 |
| TDT | Baseline | Trial data | 58.7% | 49.3% | [54.2, 63.0] | 80.6% | 39.6% | [76.8, 83.9] | 89.6% | 30.6% | [85.1, 92.8] | 28.3% | 45.1% | [23.0, 34.2] | 484 | 240 | 244 |
| TDT | RAG | Hotel reviews | 52.9% | 49.9% | [49.4, 56.3] | 96.1% | 19.3% | [94.6, 97.3] | 99.0% | 10.0% | [97.5, 99.6] | 6.8% | 25.1% | [4.7, 9.6] | 800 | 400 | 400 |
| TDT | RAG | Cheating 1 | 73.9% | 44.0% | [69.6, 77.7] | 85.5% | 35.3% | [81.9, 88.5] | 85.4% | 35.4% | [81.4, 88.6] | 13.9% | 34.8% | [7.7, 23.7] | 448 | 376 | 72 |
| TDT | RAG | Cheating 2 | 70.5% | 45.7% | [64.8, 75.7] | 71.3% | 45.3% | [65.6, 76.4] | 72.6% | 44.7% | [66.7, 77.8] | 45.0% | 51.0% | [25.8, 65.8] | 268 | 248 | 20 |
| TDT | RAG | Miami friend opinions | 50.4% | 50.0% | [46.9, 53.8] | 89.9% | 30.2% | [87.6, 91.8] | 90.2% | 29.7% | [86.9, 92.8] | 10.5% | 30.7% | [7.9, 13.9] | 800 | 400 | 400 |
| TDT | RAG | Trial data | 57.9% | 49.4% | [53.4, 62.2] | 83.1% | 37.6% | [79.5, 86.1] | 91.2% | 28.3% | [87.0, 94.2] | 25.0% | 43.4% | [20.0, 30.8] | 484 | 240 | 244 |
| VA | Baseline | Hotel reviews | 71.9% | 45.0% | [68.7, 74.9] | 57.1% | 49.5% | [53.7, 60.5] | 79.0% | 40.8% | [74.7, 82.7] | 64.8% | 47.8% | [59.9, 69.3] | 800 | 400 | 400 |
| VA | Baseline | Cheating 1 | 32.4% | 46.8% | [28.2, 36.8] | 24.3% | 43.0% | [20.6, 28.5] | 24.2% | 42.9% | [20.1, 28.8] | 75.0% | 43.6% | [63.9, 83.6] | 448 | 376 | 72 |
| VA | Baseline | Cheating 2 | 53.7% | 50.0% | [47.8, 59.6] | 50.0% | 50.1% | [44.1, 55.9] | 52.0% | 50.1% | [45.8, 58.2] | 75.0% | 44.4% | [53.1, 88.8] | 268 | 248 | 20 |
| VA | Baseline | Miami friend opinions | 52.4% | 50.0% | [48.9, 55.8] | 17.6% | 38.1% | [15.1, 20.4] | 20.0% | 40.1% | [16.4, 24.2] | 84.8% | 36.0% | [80.9, 87.9] | 800 | 400 | 400 |
| VA | Baseline | Trial data | 62.4% | 48.5% | [58.0, 66.6] | 29.8% | 45.8% | [25.9, 34.0] | 42.1% | 49.5% | [36.0, 48.4] | 82.4% | 38.2% | [77.1, 86.6] | 484 | 240 | 244 |
| VA | RAG | Hotel reviews | 65.9% | 47.4% | [62.5, 69.1] | 47.4% | 50.0% | [43.9, 50.8] | 63.2% | 48.3% | [58.4, 67.8] | 68.5% | 46.5% | [63.8, 72.9] | 800 | 400 | 400 |
| VA | RAG | Cheating 1 | 29.7% | 45.7% | [25.6, 34.1] | 20.3% | 40.3% | [16.8, 24.3] | 20.2% | 40.2% | [16.5, 24.6] | 79.2% | 40.9% | [68.4, 86.9] | 448 | 376 | 72 |
| VA | RAG | Cheating 2 | 55.6% | 49.8% | [49.6, 61.4] | 53.4% | 50.0% | [47.4, 59.2] | 54.8% | 49.9% | [48.6, 60.9] | 65.0% | 48.9% | [43.3, 81.9] | 268 | 248 | 20 |
| VA | RAG | Miami friend opinions | 50.9% | 50.0% | [47.4, 54.3] | 12.4% | 33.0% | [10.3, 14.8] | 13.2% | 33.9% | [10.3, 16.9] | 88.5% | 31.9% | [85.0, 91.3] | 800 | 400 | 400 |
| VA | RAG | Trial data | 59.7% | 49.1% | [55.3, 64.0] | 21.7% | 41.3% | [18.3, 25.6] | 31.2% | 46.4% | [25.7, 37.4] | 87.7% | 32.9% | [83.0, 91.3] | 484 | 240 | 244 |

*Note*. D&C = Details and complications; FFT = Four-Factor Theory, IDT = Interpersonal Deception Theory, IMT = Information Manipulation Theory, TDT = Truth-Default Theory, VA = Verifiability Approach. Cheating 1 = non-interactive cheating interviews; Cheating 2 = interactive cheating interviews. 95% CIs are Wilson score intervals.

**Figure 1**

*Two-Way Interaction Effect Results*

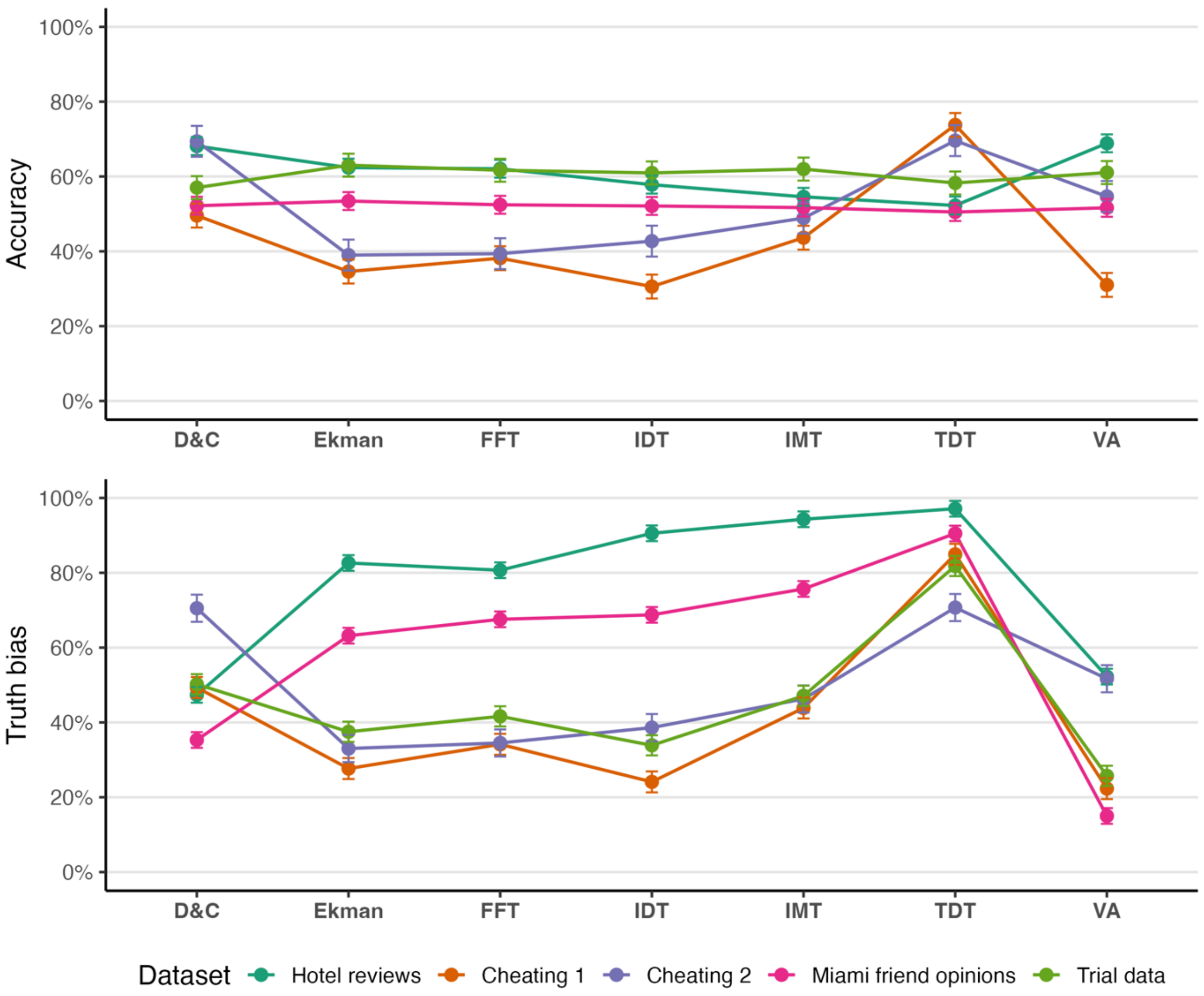

*Note*. D&C = Details and complications; FFT = Four-Factor Theory, IDT = Interpersonal Deception Theory, IMT = Information Manipulation Theory, TDT = Truth-Default Theory, VA = Verifiability Approach. Error bars are 95% Confidence Intervals. Cheating 1 = non-interactive cheating interviews; Cheating 2 = interactive cheating interviews.

**Figure 2**

*Three-Way Interaction Effect Results Predicting Truth-Bias*

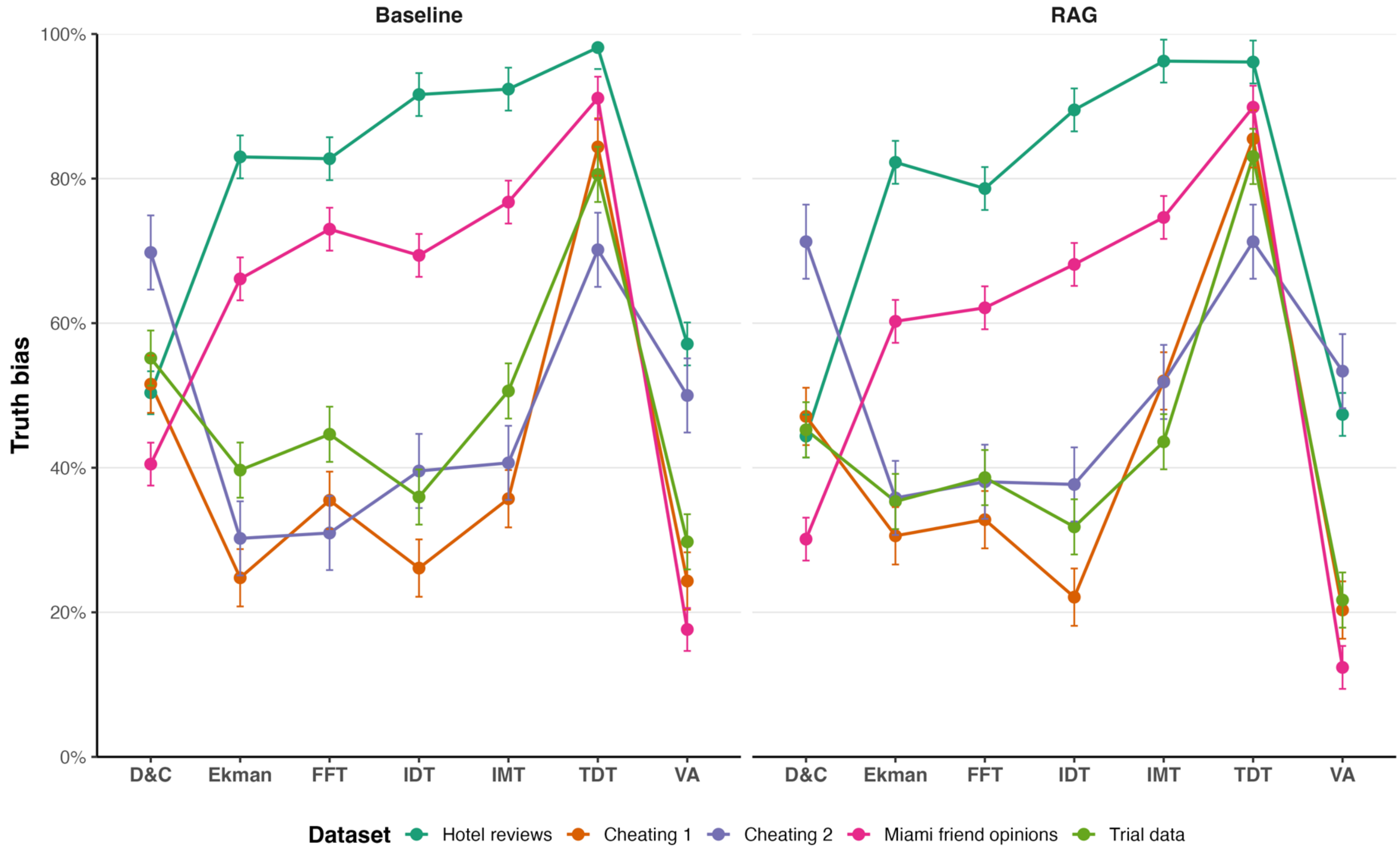

*Note*. D&C = Details and complications; FFT = Four-Factor Theory, IDT = Interpersonal Deception Theory, IMT = Information Manipulation Theory, TDT = Truth-Default Theory, VA = Verifiability Approach. Error bars are 95% Confidence Intervals. Cheating 1 = non-interactive cheating interviews; Cheating 2 = interactive cheating interviews.